\documentclass[conference]{IEEEtran}
\IEEEoverridecommandlockouts
\usepackage{cite}
\usepackage{amsmath,amssymb,amsfonts}
\usepackage{algorithmic}
\usepackage{graphicx}
\usepackage{textcomp}
\usepackage{xcolor}
\usepackage{caption} 
\usepackage{subcaption} 
\usepackage{booktabs}
\usepackage[hidelinks]{hyperref}

\def\BibTeX{{\rm B\kern-.05em{\sc i\kern-.025em b}\kern-.08em
    T\kern-.1667em\lower.7ex\hbox{E}\kern-.125emX}}
\begin{document}

\title{Balancing of competitive two-player Game Levels with Reinforcement Learning\\

\thanks{This research was supported by the Volkswagen Foundation (Project: Consequences of Artificial Intelligence on Urban Societies, Grant 98555).}
}

\author{
\IEEEauthorblockN{Florian Rupp}
\IEEEauthorblockA{\textit{Department of Informatics} \\
\textit{University of Applied Sciences Mannheim}\\
Mannheim, Germany \\
f.rupp@hs-mannheim.de} \and
\IEEEauthorblockN{Manuel Eberhardinger}
\IEEEauthorblockA{\textit{Institute of Applied Artificial Intelligence} \\
\textit{Stuttgart Media University}\\
Stuttgart, Germany \\
eberhardinger@hdm-stuttgart.de} \and
\IEEEauthorblockN{Kai Eckert}
\IEEEauthorblockA{\textit{Department of Informatics} \\
\textit{University of Applied Sciences Mannheim}\\
Mannheim, Germany \\
k.eckert@hs-mannheim.de}
}

\IEEEoverridecommandlockouts
\IEEEpubid{\makebox[\columnwidth]{979-8-3503-2277-4/ 23/\$31.00 ©2023 IEEE \hfill}
\hspace{\columnsep}\makebox[\columnwidth]{ }}

\maketitle

\IEEEpubidadjcol

\begin{abstract}
The balancing process for game levels in a competitive two-player context involves a lot of manual work and testing,
particularly in non-symmetrical game levels.
In this paper, we propose an architecture for automated balancing of tile-based levels within the recently introduced PCGRL framework (procedural content generation via reinforcement learning).

Our architecture is divided into three parts: (1) a level generator, (2) a balancing agent and, (3) a reward modeling simulation. By playing the level in a simulation repeatedly, the balancing agent is rewarded for modifying it towards the same win rates for all players. To this end, we introduce a novel family of swap-based representations to increase robustness towards playability. We show that this approach is capable to teach an agent how to alter a level for balancing better and faster than plain PCGRL. In addition, by analyzing the agent's swapping behavior, we can draw conclusions about which tile types influence the balancing most. We test and show our results using the Neural MMO (NMMO) environment in a competitive two-player setting. 
\end{abstract}




\section{Introduction}

The design of levels is a key concept when creating games. To keep players engaged, the balance between a challenging and enjoyable experience must be found. This is generally not an easy task, as it also depends on the skill and experience of the players.
Furthermore, game levels for competitive multi-player games must be designed to be balanced towards equal initial win chances for all players. Imbalanced games will result in boredom or frustration and players will quit playing~\cite{andrade_dynamic_2006, becker_what_2020}.

To ensure balancing through the level design, game designers often rely on almost (point) symmetric map architectures in many games. This can be seen in popular competitive esports titles such as League of Legends, DotA~2 or Starcraft~2, but also in other competitive tile-based games such as Advance Wars or Bomberman.
Symmetrical levels, however, may not be the only way to go. Non-symmetrical levels offer more diversity and can create new ways for playful creativity to be entertaining and challenging.

Several approaches have been proposed using procedural content generation (PCG) for level balancing in previous works, such as search-based approaches~\cite{togelius_search-based_2011,sorochan_generating_2022}, evolutionary algorithms~\cite{lanzi_evolving_2014,lara-cabrera_balance_2014,karavolos_using_2018} or graph grammars~\cite{kowalski_strategic_2018}.
In this work, we introduce the balancing of tile-based levels with reinforcement learning (RL). Once a RL model is trained, it can generate levels fast and is less dependent on randomness compared to evolutionary methods. In addition, we develop our method to be not dependent on a single domain, as is the case with graph grammars or search-based approaches.
Therefore, we use the recently introduced PCG for RL framework (PCGRL)~\cite{khalifa_pcgrl_2020}.

\begin{figure}
     \centering
        \includegraphics[width=2.8in]{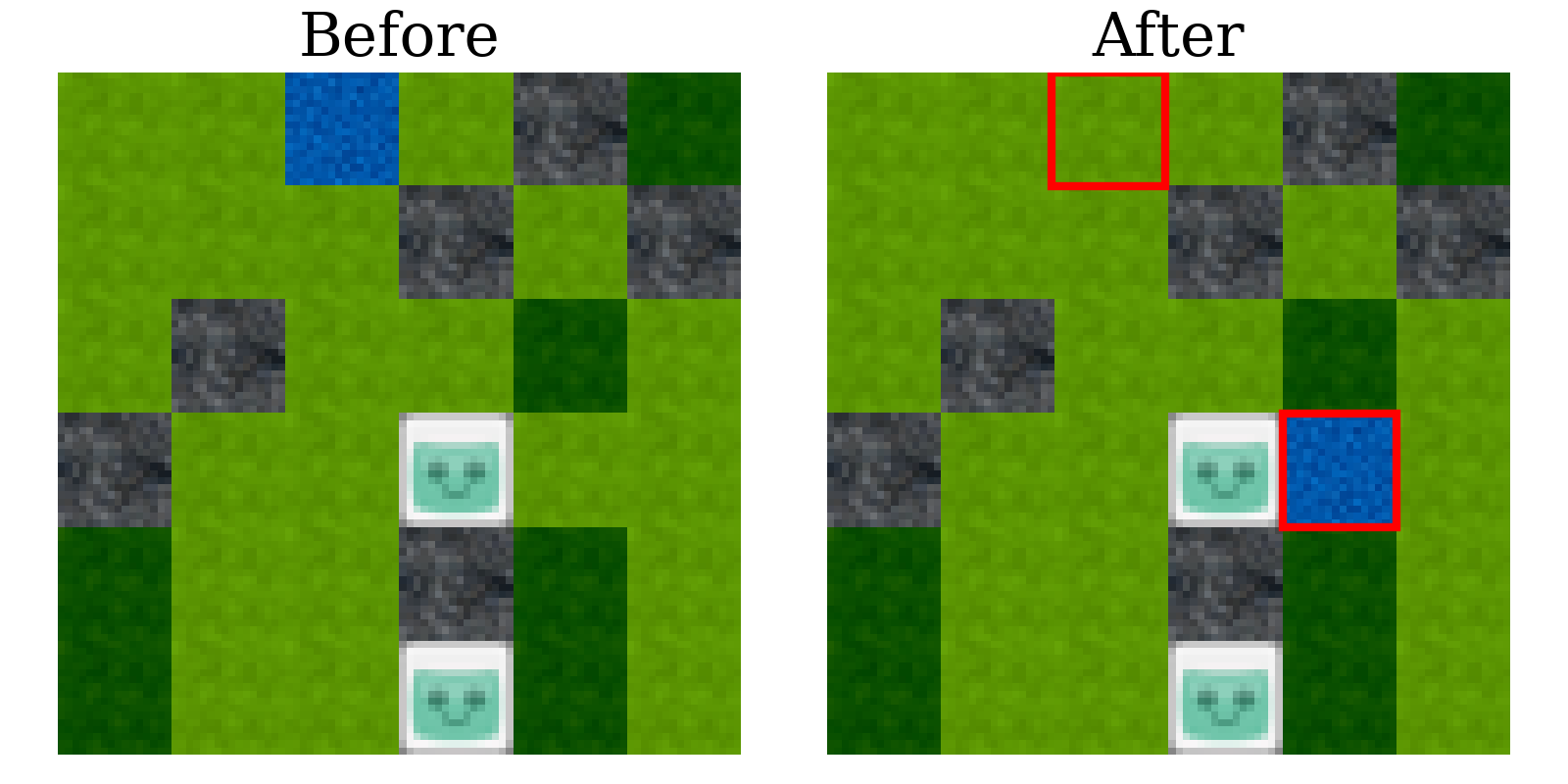}
    \caption{
In our game environment, two players must forage for resources like food (dark green) and water (blue) to survive longest. By swapping the highlighted tiles, the trained model reached a more balanced game in simulated game runs. 
}
     \label{fig:gen-level1}
\end{figure}

Our method is evaluated on the openly available Neural Massively Multiplayer Online (NMMO) environment~\cite{suarez_neural_2019} which has been designed for competitive multi-player research. We frame the problem as a resource gathering and survival game. Figure~\ref{fig:gen-level1} shows an example of how our method balanced a generated level by swapping highlighted tiles.

Our contributions are:
\begin{itemize}
    \item A domain independent architecture for learning the automated altering of game levels towards equal balancing with RL. 
    \item A novel swap-based representation pattern to frame the problem as Markov decision process.
    \item An experimental study to evaluate the proposed architecture compared to the original PCGRL method.
\end{itemize}
This paper is structured as follows: We give a brief overview of related work~(\ref{sec:related-work}) and the background~(\ref{sec:background}). In Section~\ref{sec:method} the method is described and the implementation details in~\ref{sec:details} subsequently. Experiments and results are presented in~\ref{sec:experiments}, followed by the discussion~(\ref{sec:discussion}) and conclusion~(\ref{sec:conclusion}).

\section{Related work}
\label{sec:related-work}
For multiplayer \emph{game balancing}, several methods have been proposed in the past. One approach is the balancing through the configuration of game entities such as characters, weapons, items etc. To do this, the games industry uses pipelines to extract data-driven insights from played games to adjust the balancing in a regular schedule\footnote{https://www.leagueoflegends.com/en-gb/news/dev/dev-balance-framework-update/}. Another approach trains models with data from played games to replicate human behavior for automated balancing of game entities~\cite{pfau_dungeons_2020}. In~\cite{sorochan_generating_2022} game entities are generated and balanced with a search-based method.

A different approach, especially in online games, are matchmaking algorithms. These methods assign which players play against each other or are in the same team. This is a contribution to balancing by ensuring that players' skill in a match is at a comparable level. Multiple works on this exist, however, many are based on a form of the Elo rating system~\cite{elo_rating_1978} used by the World Chess Foundation.

The balancing of a level is furthermore influenced by its design. That refers to e.g., the position where game entities are located in relation to players' initial spawn position. As aforementioned, it is ensured with (point) symmetric map architectures in many games.
Contrary to the intended balancing the perception of players may differ. A survey concluded that a prevailing opinion among high experienced players is that only symmetrical levels can actually be balanced~\cite{togelius_controllable_2013}.

To automatically construct balanced levels with PCG, work has been introduced using evolutionary algorithms~\cite{lanzi_evolving_2014,lara-cabrera_balance_2014}. In contrast to this work, the fitness for balancing is evaluated rule based with heuristics or metrics. These metrics are either handcrafted or rely on a data-driven method using played games. In~\cite{karavolos_using_2018} this is extended with a convolutional neural network to predict game related information which are then used in the fitness function of an evolutionary algorithm to produce balanced levels. A different approach is the usage of graph grammars~\cite{kowalski_strategic_2018}. The balancing here is constructed by the rule-based placing of strategic game entities, which is, however, very dependent on the domain.


In this work, we focus on \emph{procedural content generation} for level generation, but the term PCG is also used in other applications such as the generation of character models or textures. PCG methods span over from dedicated algorithms~\cite{mojang_mincraft_2011}, search based methods~\cite{togelius_search-based_2011} up to the use of machine learning~\cite{summerville_procedural_2018} and deep learning~\cite{liu_deep_2021, giacomello_doom_2018, awiszus_toad-gan_2020, awiszus_world-gan_2021}. 
Once trained, machine learning-based methods can quickly generate content as needed, but they rely on having game levels already available from which a model can be learned. Therefore, when creating new games, these methods are usually not applicable efficiently.

In the games community, RL has been widely applied to \emph{play} games~\cite{silver_general_2018, vinyals_grandmaster_2019}, but recently, it has also been used for PCG~\cite{khalifa_pcgrl_2020, bontrager_learning_2021, gisslen_adversarial_2021}.
There are already quite a number of methods which adapted the PCGRL method. In \cite{earle_learning_2021} controllable content generators were introduced, where users control the generated content with additional constraints, such as e.g., the number of players.  
We do, however, not compare our method with it, as our objective is a balanced level only.
Other approaches use evolutionary strategies on top of the PCGRL framework to achieve more content diversity \cite{khalifa_mutation_2022} or adapted the method for 3D environments \cite{jiang_learning_2022-4}.

\section{Background}
\label{sec:background}
\subsection{The PCGRL framework}
In this work, we use the PCGRL \cite{khalifa_pcgrl_2020} framework for level balancing.
In PCGRL, PCG is formulated as a sequential decision-making task to maximize a given reward function, where semantic constraints can be expressed and thus, no training data is needed. To apply RL, the PCG problem is framed as a Markov decision process (MDP).

Therefore, PCGRL introduces three different MDP representations for level generation. 
We will introduce three new swap representations based on the existing representations. Later we argument that they are a better fit to alter game levels to improve e.g., the balancing state. The original representations in PCGRL are: 

\paragraph{Narrow} This representation randomly selects a tile in the grid and the agent only needs to decide what type of tile should be placed on the selected position. This has a small action space since it only consists of the different types of tiles. 

\paragraph{Turtle} The \emph{turtle} representation allows the agent to move on the map and then to decide what type of tile to put on the current position. The advantage to the narrow representation is that the agent is not restricted to the randomly assigned position and therefore can learn where to move next.

\paragraph{Wide} In the \emph{wide} representation the agent is given full control of the level generation process as the agent can decide which tile of the whole grid should be changed. This increases the action space greatly as every position of the grid multiplied by the number of tiles represents an action. This is the most human-like representation as the agent can change everything directly according to a plan it has constructed.

\subsection{The Neural MMO Environment}
\label{sec:nmmo}
As basis for our experiments, we use the NMMO environment~\cite{suarez_neural_2019}. The NMMO environment is a Massively Multiplayer Online Battle Royal inspired multi agent RL environment for research.
It simulates a tile based virtual world where agents must forage for resources to survive and be prepared for fights with other agents. All actions of the agents are processed simultaneously per time step. The agent which survives the longest wins the game.
The agent's state is represented by its position as well as its current health, food, and water level.

\begin{figure}
     \centering
     \includegraphics[width=0.7\linewidth]{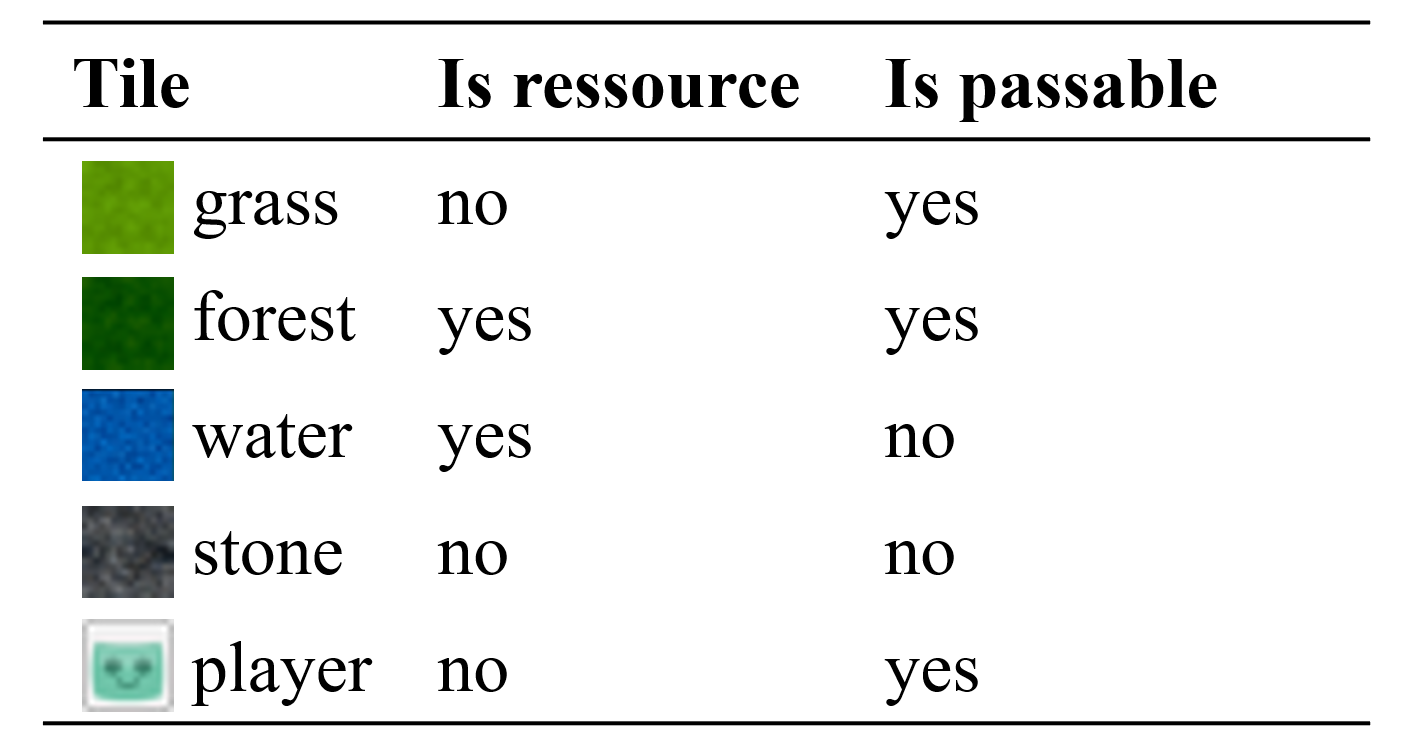}
     \caption{Description of the NMMO tiles.}
     \label{fig:tile-descr}
\end{figure}

In this paper, we focus on the balancing in the PCG process of the game map. To apply and evaluate our method, we simplify the win condition and frame the game as a forage survival game. Therefore, we disable the combat system and limit map tiles to the types of grass, forest, stone, and water. The agent cannot move on stone nor water tiles. Forest and water are resources the agent can gather. A detailed overview of the tiles and their behaviors is given in Figure~\ref{fig:tile-descr}. When moving on a forest tile, it is consumed automatically as food and refills the agent's food indicator. In the process the forest tile changes to the state \emph{scrub}. Scrub tiles cannot be consumed, however, there is a probability of 2.5\% per time step to respawn and transit back to the forest state.
To refill the agent's water indicator it must simply step on an adjacent tile. Water tiles are not depleted. If one of both resource indicators are empty the agent loses a fixed amount of health per step. If the agent's health indicator drops to zero, the agent dies. An agent can restore health having water and food indicators over 50\%. 

For this research we frame the win condition equally for both agents. The agent which completes one of two goals first wins:
\begin{itemize}
    \item Collect five food resources.
    \item Last agent standing: Not starved nor died of thirst.
\end{itemize}

In this introductory research, we fix the map size to a grid of 6x6 tiles and two players to speed up the experiments and to keep the results interpretable by human evaluators. Furthermore, the smaller grid size limits the available space and resources which reinforces the competitive race of the agents. This competitive race is to be balanced in this paper.

\section{Method}
\label{sec:method}
\subsection{Balancing architecture}
\label{sec:architecture}

\begin{figure*}
  \centering
  \includegraphics[width=0.9\linewidth]{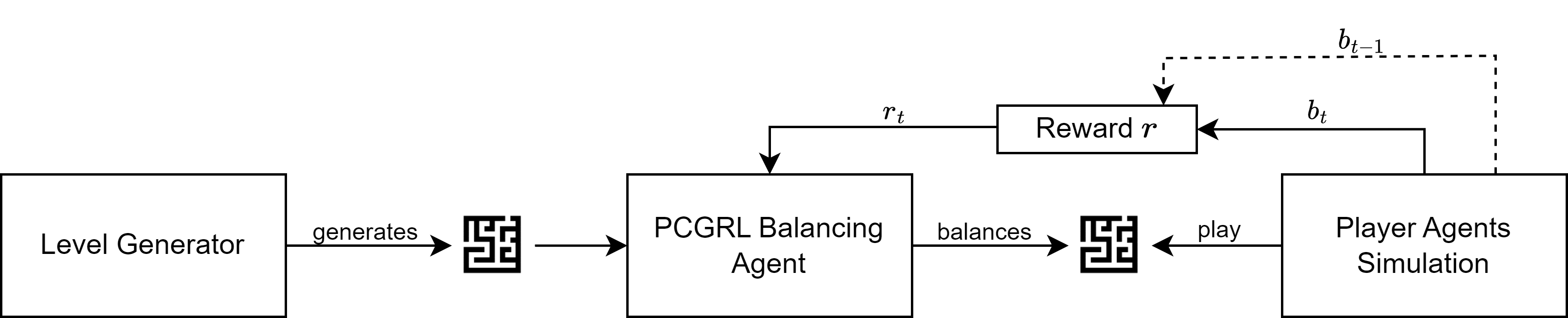}
  \caption{Description of the balancing architecture. It is separated into three units: A level generator, a level balancing agent and a level playing simulation. In the latter, the game is simulated by playing it $n$-times with player agents. The reward $r_t$ for training the balancing agent is calculated out of the balancing states $b_t$ and $b_{t-1}$ from the simulations.}
  \label{fig:architecture}
\end{figure*}

This paper proposes an architecture to balance tile-based levels with RL. The idea is to reward a PCGRL agent for only altering a given playable level towards a balanced level.
A balanced level is a level where all agents with the same skill have the same win rate.
The architecture is depicted in Figure~\ref{fig:architecture}. It is separated into three parts: a level generator, a balancing agent and a playing simulation. 

The level generator constructs a playable level from random noise which is then fed to the balancing agent. Here we use PCGRL for the generator, however, other PCG methods will serve at this point comparable. The generator is trained separately before training the balancing agent.

The core idea to balance the level is not to generate a new level but to modify the given level accordingly to fulfill the balancing constraint.
Each time step the balancing agent can decide to swap the positions of two selected tiles. If a swap (Section \ref{sec:swap}) was made, the level is played $n$-times in a simulation by player agents. Subsequently, the balancing agent is rewarded based on how this action influenced the balancing state in simulations (Section \ref{sec:reward}). This is inspired from the idea of simulation-based fitness functions in~\cite{togelius_search-based_2011}.

More details are given in the implementation details~(Section~\ref{sec:details}).

\subsection{Swap-based representation pattern}
\label{sec:swap}
To formulate the problem of PCG with RL as MDP three representations have been introduced in \cite{khalifa_pcgrl_2020}. In all of them the agent can decide to replace a tile at a distinct position. This method, however, may result in unplayable levels at a time step. Furthermore, to move the position of e.g., a player tile somewhere else, the agent would need to remove the player tile first before creating it at a different position. In this time step the level would not be playable for the player agents and thus, no reward could be computed in the simulation step. Additionally, the agent would get a negative reward first due to the number of players is now invalid. The subsequent creation at a different position under the previously given negative reward is hard to learn for RL agents.

For this reason, we introduce a swapping-based representation pattern. In these representations the agent can decide to swap the positions of two tiles per time step.
Not adding or removing tiles entirely is a more robust approach to ensure level playability.

There might be, however, game domains where multiple tiles have a semantic connection e.g., multiple river tiles construct a river. Swapping these tiles around can break playability. For this reason, we suggest in the balancing step to feed unplayable levels back to the generator to fix it again. This repair is similar to the level fixing as in~\cite{siper_path_2022}. In this work we demonstrate the power of swapping with a simpler domain where no semantic relations of multiple tiles exist. Also, just swapping tiles ensures that the old level isn't simply generated from scratch since that's the job of the level generator.
Swapping two tiles of the same type has no effect on balancing. Thus, we prevent these swaps to reduce computational intensity. In these cases, the agent is rewarded with 0.

Therefore, we extend the representations narrow, turtle and wide~\cite{khalifa_pcgrl_2020} with a swapping mechanism:
\paragraph{Swap-Narrow} As in PCGRL the agent observes the tiles in the environment one-hot encoded. At each time step two random tile positions are presented to the agent, and it can decide to swap the tiles or not. The agent's little position control results in in a very small action space $A$ with only two actions: swap or do not swap. $A$ is therefore $A=[2]$.
\paragraph{Swap-Turtle} Starting at two random positions the agent can decide to swap the tiles at the current positions in each time step. If no change is made it can decide to which adjacent tile to move next. The observations are in a one-hot encoded format as well. $A$ is therefore $A=[4,4,2]$ which results in 32 possible actions.
\paragraph{Swap-Wide} In this representation the agent sees the whole level and can freely determine the tile positions and whether to swap it. The observations are provided one-hot encoded. It can be interpreted as looking at the whole level and then decide what to move where. A drawback here is the large action space since it scales twice with the grid width $w$ and height $h$. $A$ is therefore $A=[w,h,w,h,2]$. In the case of a square grid shape as in this work with a size of 6, $A$ has 2592 possible actions.

\subsection{Reward design}
\label{sec:reward}
For the successful training of a RL agent the reward function is crucial. Heuristic approaches for the evaluation of balancing have been introduced in~\cite{lanzi_evolving_2014,lara-cabrera_balance_2014}. These approaches, however, include domain specific information and can thus not be transferred. To address this shortcoming, we propose the use of a more generic, game domain independent reward function only relying on how often one player wins.

The reward is designed to reward the agent for improving the balancing state $b_t$ in time step $t$ in comparison to $b_{t-1}$. $b_t$ is calculated using the results from the player agent simulation step. In this step we let the player agents play the game level $n$-times. The win conditions must be designed to have at least one winner. We implement a reward function that can handle draws as well, thus draws indicate the level was balanced towards both sides.

For the calculation of the balancing state $b_t$ we assign indices to the two players first where $p_1=0$ and $p_2=1$. For all $n$ simulations we count which player(s) won and save their player indices as winners $w_t$. The number of winners in the simulations are denoted as $wl_t$. Since draws are permitted, $n \le w_l \le 2n$. We calculate $b_t$ using Equation~\ref{eq:bt}.

\begin{equation}
\label{eq:bt}
b_t = \frac{1}{wl_t} \sum^{w_t}_{i=0} w_{t,i}
\end{equation}

$b_t$ is defined in $[0,1]$ where 0.5 indicates both players won the same number of games, so the balance is maximal. The values 0 and 1 indicate a maximal unbalanced level where one distinct player wins every game. 0 indicates player 1 is only winning, 1 player 2. The reward $r_t$ is then: $ r_t = b_{t-1} - b_t  + \alpha$.

To reward the agent for maximum balancing, an additional reward $\alpha$ is given if $b_t$ is exactly 0.5. Otherwise, $\alpha$ is 0. As a result, the reward will be positive if the agent improves the balancing state, negative otherwise. For no impact on the balancing state the agent receives no reward (value 0).

\section{Implementation details}
\label{sec:details}
\subsection{Level generator}
\label{sec:gen}

The task of this unit is to generate playable levels for balancing. In this work, the generator is a model trained using the PCGRL framework.

The reward for the training process is designed to reward the agent to
\begin{itemize}
    \item have exactly two players and
    \item create a valid path between both players.
\end{itemize}

To guarantee the direct competition, the latter constraint ensures that both players have access to the same area of the game level. Additionally, this prevents single players being locked within stone walls.

The NMMO environment uses by default PCG to generate random levels, however, the players' spawn positions are randomly determined when the game starts. Nonetheless, the spawn positions are crucial for balancing and thus, should be possible to be aware of and to influence by the balancing agent. For this reason, a player tile is integrated in the level generation process representing the positions.

If a valid path between both players exists the agent receives a positive reward, a negative otherwise. Additionally, each step the agent is rewarded with the difference of players there are and should be (similar like in \cite{khalifa_pcgrl_2020}). We achieve the best results using the \emph{wide} representation. An episode ends when both constraints are met, or the agent exceeds a fixed number of permitted steps or changes.

After training, the model can produce levels satisfying the given constraints in 98,7\% levels out of an evaluation sample of 5000. The diversity of generated levels is at the maximum at 100\%. An example of a generated level is shown in Figure~\ref{fig:gen-level1}~(left) and a tile legend is given in Figure~\ref{fig:tile-descr}.

The yielded levels are then further proceeded to the balancing agent. Level generation at this step is done until the agent produced a valid level fulfilling the constraints to ensure the balancing agent receives playable levels only.


\subsection{Balancing agent}
\label{sec:balancing}
The balancing agent is the core component of the architecture in Figure \ref{fig:architecture}. We model the agent as PCGRL agent which can decide to alter a previously generated game level per time step. To support this process, we use swapping representations.
The reward is calculated out of the results of a simulation of the balancing state (Section \ref{sec:reward}).

The observation is the current level one-hot encoded. In detail, this depends on the chosen representation. As RL algorithm, we apply PPO (proximal policy optimization)~\cite{schulman_proximal_2017}. An episode ends when the level is either balanced, or a fixed number of steps or changes is exceeded.

\subsection{Player simulation}
To calculate $b_t$ for the reward (Section~\ref{sec:balancing}) we run a simulation $n$-times of player agents playing the game.
The player agents can be any solution which can simulate a player's behavior at a required quality. In this work we use the scripted \emph{Forage}-agent for the NMMO environment which is publicly available\footnote{https://github.com/NeuralMMO/baselines}. The usage of scripted agents provides the benefit of a deterministic behavior at any time step. If the agent's water or food indicator drops below a certain threshold it looks for the next available resource of that type to refill it.
At this point, however, the application of trained agents with e.g., RL, different skilled or types of agents is feasible as well. See the discussion~(Section~\ref{sec:discussion}) for further information.

Despite running the simulation with the same player agents, winners may differ each pass. This is because of probabilistic mechanics making games interesting such as rolling dices or the respawn of resources at a certain probability. A player may win with luck, but when playing many times, skill should make the difference. The question is then how often must a simulation be run to minimize the noise of probability? Due to running simulations is computationally intensive, it is of interest to find the number of minimal runs where win rates vary at an acceptable level.

We approach this by investigating how much the win rates $w_n$ of a specific number of simulations $n$ differ on average from $n-2$. Only even values for $n$ are applicable since otherwise a balanced game is not possible. Therefore, we run the simulation on a sample of size $s=500$ levels up to a number of 30 times and compare how win rates for $n$ fluctuate on average in comparison to $n-2$. The average deviation~$\mu_n$ of a distinct $n$ is expressed with: $\mu_n = \frac{1}{s} \sum^{N}_{n=0} | w_n - w_{n-2} |$.

\begin{figure}[!t]
  \centering
  \includegraphics[width=3.1in]{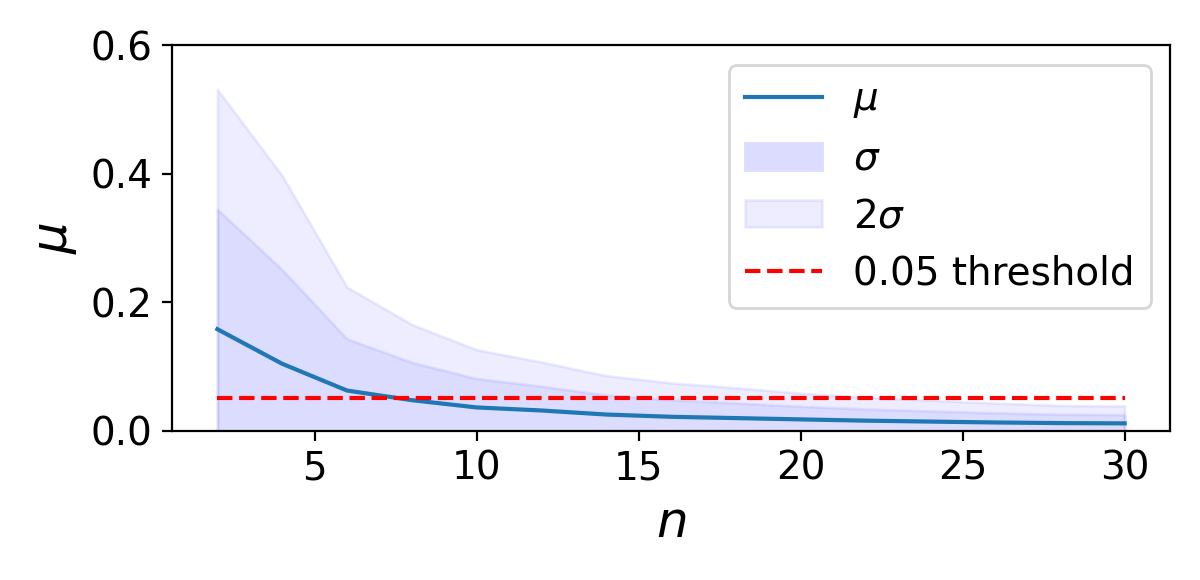}
  \caption{How many times $n$ should we run the simulation to approximate the balancing state? We figure that out by calculating the mean deviation $\mu_n$ of win rates from $n-2$ to $n$ for the investigation of fluctuations in win rates.}
  \label{img:balancing}
\end{figure}

This ratio is depicted in Figure~\ref{img:balancing}. In addition, the standard deviation $\sigma$ and $2\sigma$ is given. It is clear that the larger $n$ is, the smaller $\mu_n$ is. To determine a suitable $n$ we set the threshold: $\mu_n + \sigma < 0.05$. For this game with the configured player agents this is true for $n \ge 14$. Thus, we use 14 as $n$ in this work.

This method can be used to determine $n$ for any domain.

\section{Experiments and Results}
\label{sec:experiments}

We evaluate our architecture in several steps: First of all, we sample a fixed data set of levels to use for direct comparison with the generator. Then we compare the balancing performances of the three introduced representations with the original PCGRL method as a baseline~(\ref{sec:eval-swap}). To generate balanced levels with the original PCGRL method, we integrated the balancing constraint into the reward function.

Subsequently, we investigate on the levels the models created~(\ref{sec:gen-levels}) and examine which tiles the models swapped in the level altering process~(\ref{sec:impact}). The latter can provide insights about which tiles actually influences the balancing.

\subsection{Performance overall}
\label{sec:eval-swap}
We compare the performance of the swap representations among themselves and in contrast to the original PCGRL as baseline\footnote{Only the narrow representation is shown here; the other two yield similar results.}. For direct comparison all models have been trained the same amount of 200k training steps resulting in 970 updates of the policy.

\begin{figure}[!t]
  \centering
  \includegraphics[width=2.9in]{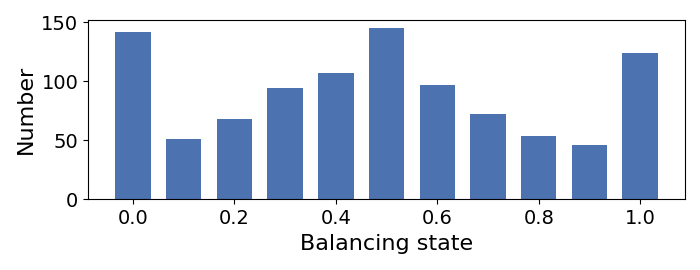}
  \caption{Distribution of the initial balancing states in the generated data set of 1000 levels. We use this data set to compare the different representations.}
  \label{fig:init-balancing}
\end{figure}

For evaluation we create a data set of 1000 levels with the generator. This data set is then used for all four models for evaluation.
First of all, we investigate on the distribution of initial balancing states of the levels in the data set (see Figure~\ref{fig:init-balancing}). Important to mention here is the non-equal distribution. Except the states 0 and 1 the levels seem to be normal distributed around the most balanced state of 0.5. A share of 13.6\% of levels is initially balanced. However, peaks towards maximally unbalanced levels at the outer edges of the distribution can be observed. Maximally unbalanced levels make up 26.6\% in the data set.

To evaluate the performance of the balancing method we compare the balancing state before and after the balancing per representation. A general overview is given in Table~\ref{tab-overview}. A histogram of the balancing improvement is depicted in Figure~\ref{fig:repr-compare}. 

\begin{figure}
     \centering
     \begin{subfigure}[b]{0.5\textwidth}
          \centering
          \includegraphics[width=\textwidth]{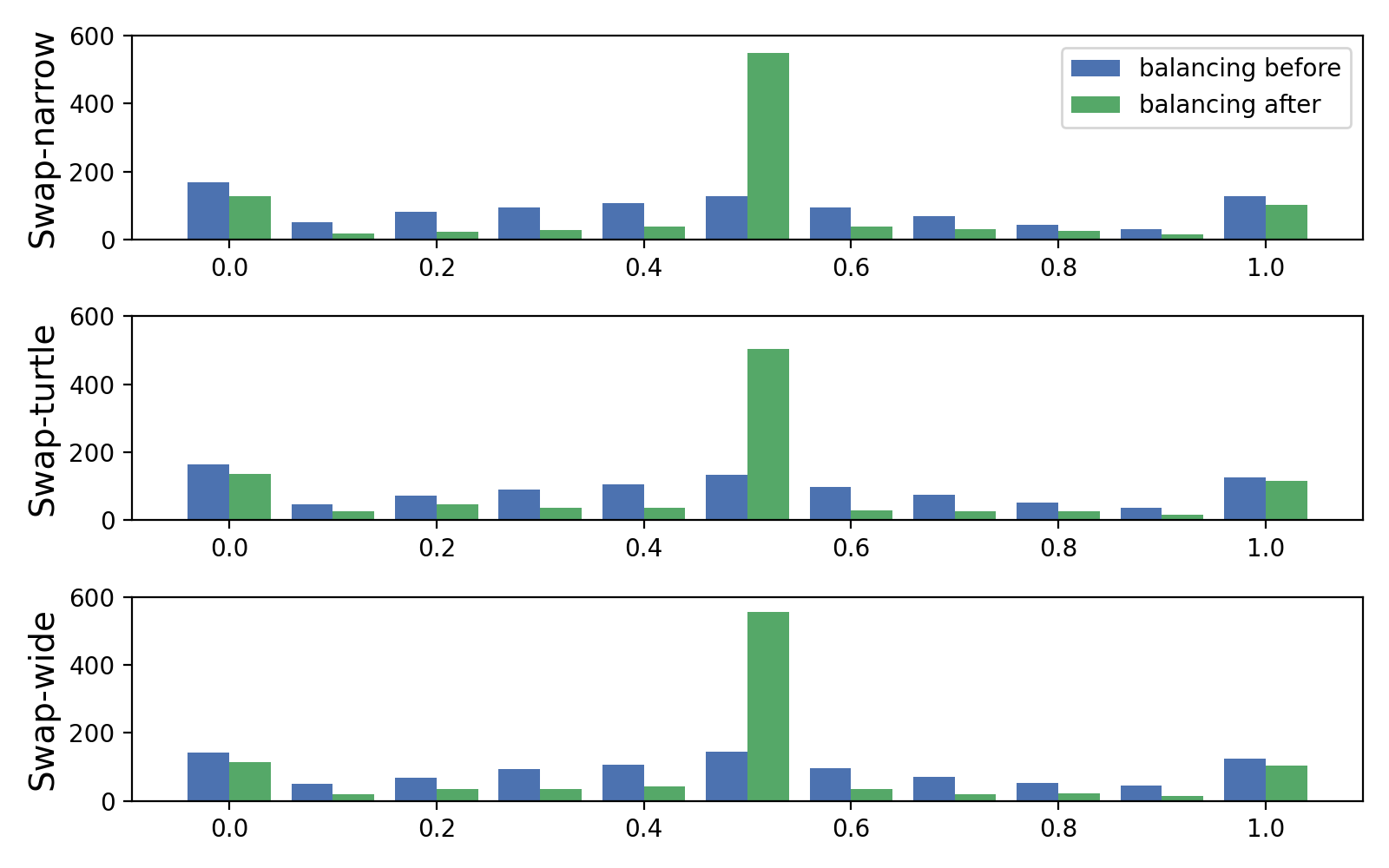}
          \caption{Swap representations}
          \label{fig:repr-compare}
     \end{subfigure}
     \vspace{\fill}
     \begin{subfigure}[b]{0.5\textwidth}
         \centering
         \includegraphics[width=\textwidth]{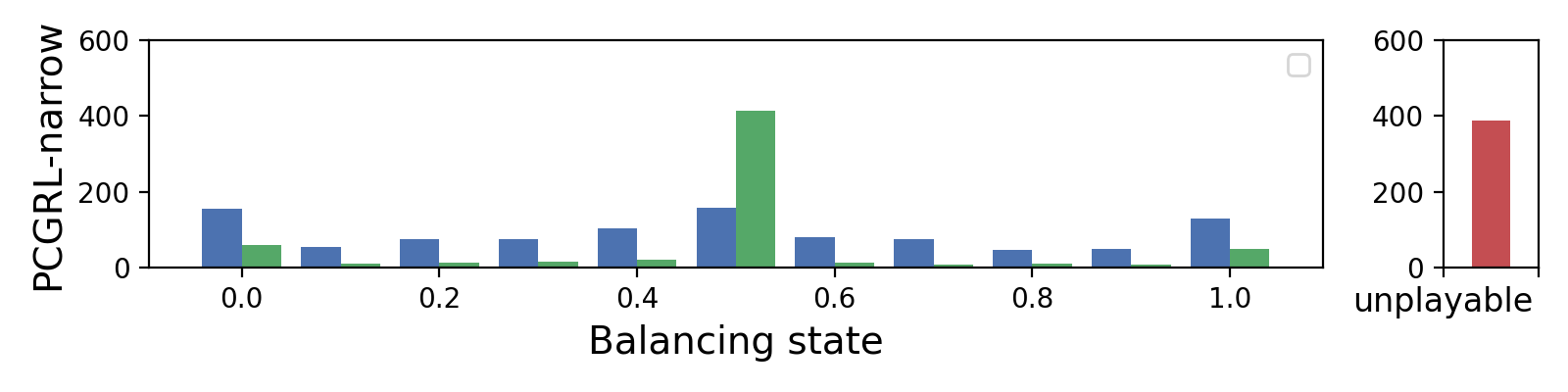}
         \caption{Original PCGRL (narrow)}
         \label{fig:narrow-pcgrl}
     \end{subfigure}
    
          \caption{Comparison of balancing state distributions before and after the balancing process per each representation. The same data set of 1000 levels is used each time. The swap representations (a) are compared to the original PCGRL implementation directly (b).}
     \label{fig:repr-compare}
\end{figure}

The performance of all three swap representations is at comparable quality. Each representation managed to improve the share of balanced levels significantly. The swap-narrow and swap-wide representations perform slightly better towards balancing in contrast to swap-turtle.
The plain PCGRL narrow representation also improved the share of balanced levels, however, with 36.7\% this share is smaller compared to the swap representations. Additionally, a proportion of 38.8\% of the levels are in an unplayable state at episode end~(Figure~\ref{fig:narrow-pcgrl}); whereas our methods result in 100\% of playable levels in all cases on this domain.
In all results, however, remain unbalanced levels. The biggest share remains for the balancing states 0 and 1.


\begin{table}[htbp]
\caption{Performance overview of the representations. Initially balanced levels were not taken into account.}
\begin{center}
\begin{tabular}{|c|c|c|c|c|}
\hline
                        & {\textbf{S-narrow$^{\mathrm{*}}$}}  & {\textbf{S-turtle$^{\mathrm{*}}$}} & {\textbf{S-wide$^{\mathrm{*}}$}} & {\textbf{PCGRL}}   \\
\hline
Balanced (\%)           & \textbf{48.1}   & 42.5          & \textbf{48.1}  & 30.4    \\ \hline
Improved (\%)           & 63.3         & 56.8          & \textbf{64.1}    & 36.7      \\ \hline
Avg. changes            & 4.6$\pm$1.9  & 4.7$\pm$1.9   & 4.9$\pm$1.9  & 5$\pm$1.8    \\ \hline
Avg. ep. length         & 11.3$\pm$6.1 & 25.4$\pm$17.7 & 14.1$\pm$7.9 & 15$\pm$7.9   \\ \hline
Size action space            & 2            & 32            & 2592         & 10           \\
\hline
\multicolumn{4}{l}{$^{\mathrm{*}}$Swap representations are abbreviated with S.}
\end{tabular}
\label{tab-overview}
\end{center}
\end{table}

\subsection{Generated levels}
\label{sec:gen-levels}
\begin{figure}
     \centering
     \begin{subfigure}[b]{0.41\textwidth}
         \centering
         \includegraphics[width=58mm]{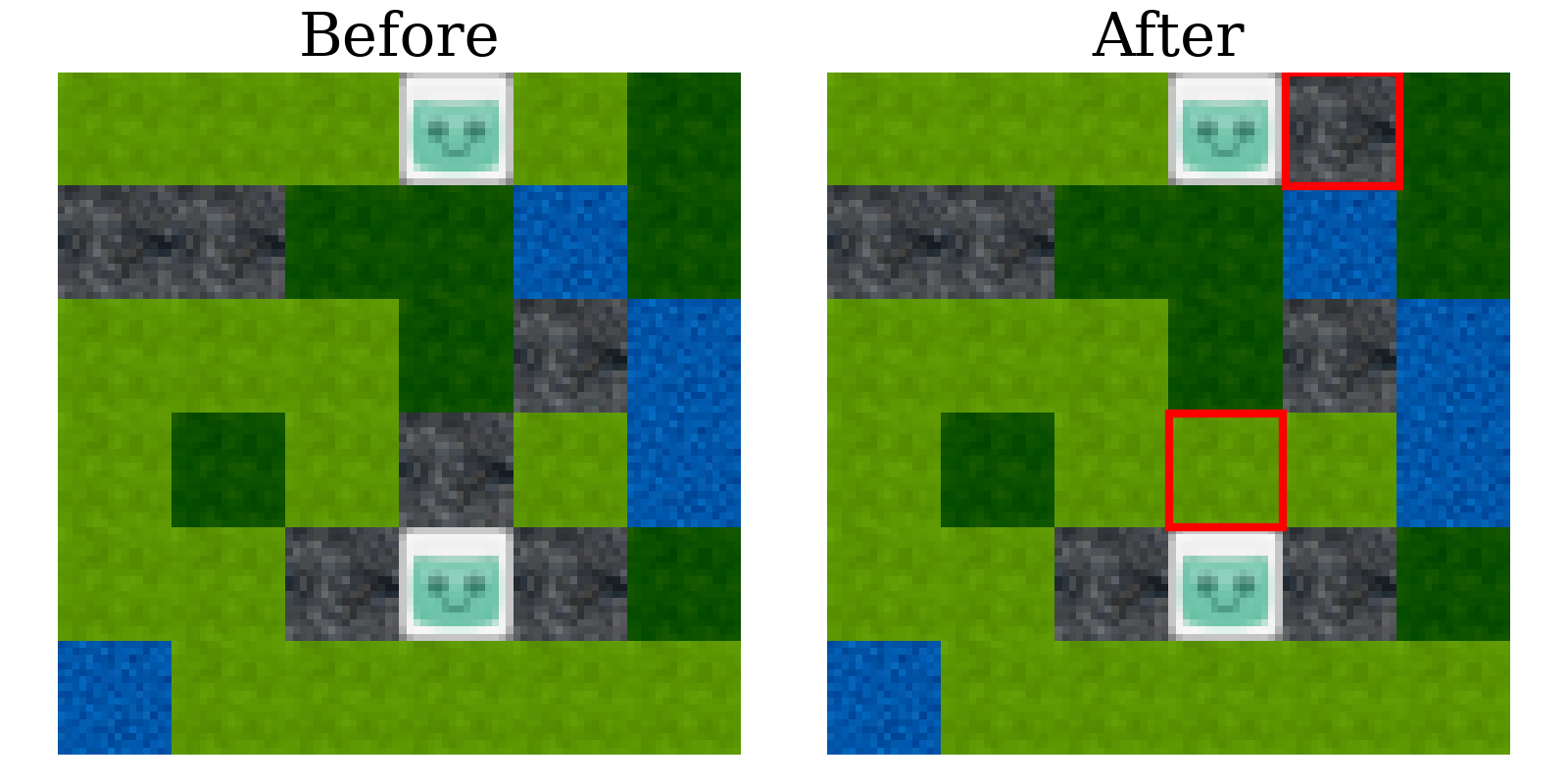}
         \caption{The agent balanced a maximally unbalanced level (0) towards a balanced level (0.5) by swapping one tile.}
         \label{fig:res-perfect}
     \end{subfigure}
      \vspace{\fill}
     \begin{subfigure}[b]{0.41\textwidth}
         \centering
         \includegraphics[width=58mm]{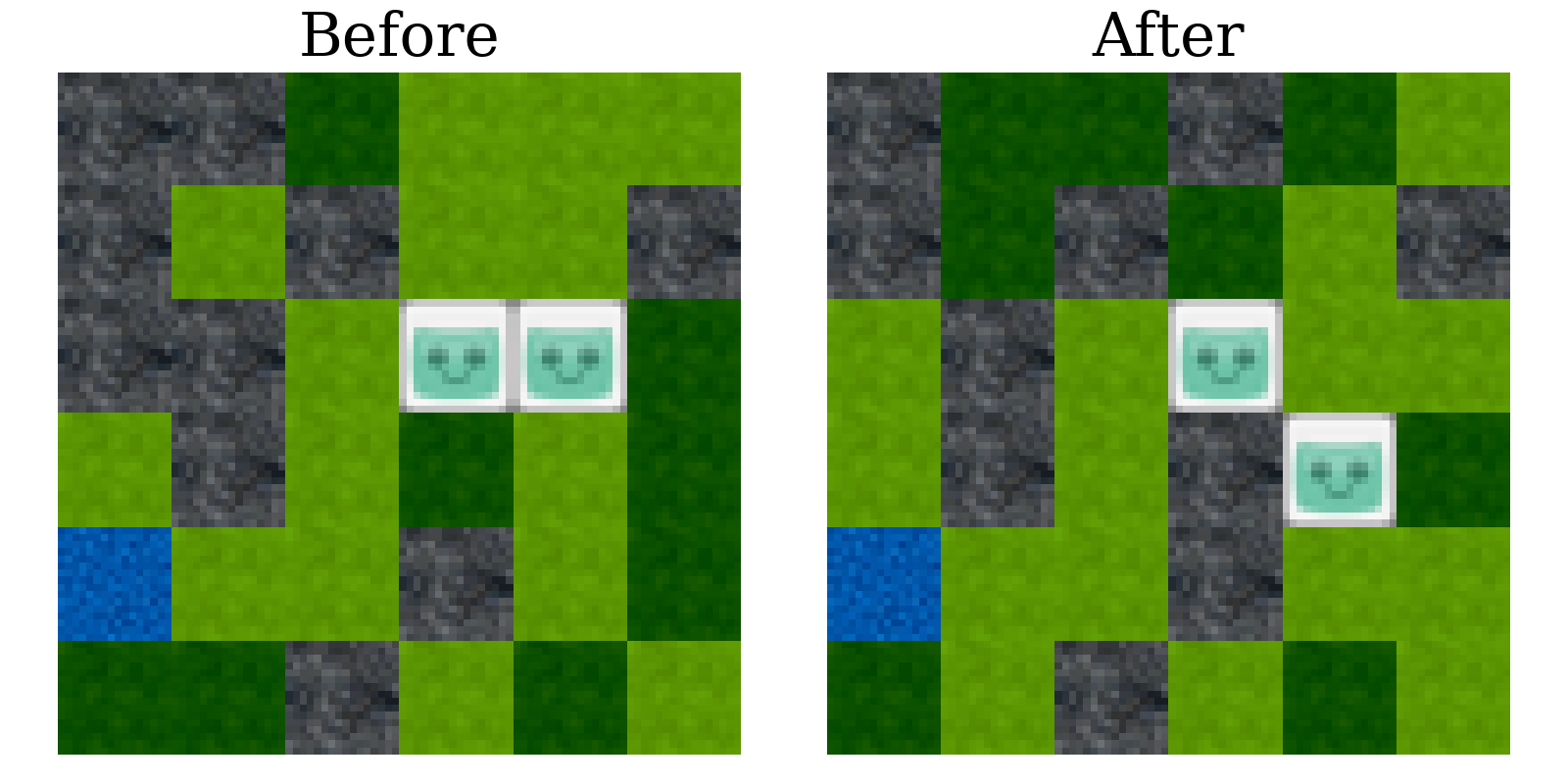}
         \caption{The agent could not change the initial balancing of 0. Generation stopped after a permitted number of changes has been exceeded. Since several tiles have been swapped, they are not highlighted for the sake of visualization.}
         \label{fig:res-failed}
     \end{subfigure}
    
    \caption{Examples of altered levels by the balancing agent. The left column shows generated levels before balancing. The right level is the result after the balancing process. Swapped tiles are indicated by red frames.}
     \label{fig:gen-levels}
\end{figure}

Figure \ref{fig:gen-level1} and \ref{fig:gen-levels} give examples of the different types of generated levels taken from the samples in Section \ref{sec:eval-swap}. Figure \ref{fig:res-perfect} is an example where the balancing agent altered the given level to a balanced level by swapping only one tile. By swapping the highlighted grass tile with the rock tile the path to the resource (forest) tiles is now blocked. This results in a more equal availability of food resources for both players. In Figure \ref{fig:gen-level1} the agent balanced the level from an initial balancing state of 0.3 to 0.5. By swapping the marked water tile to a more central position for both players the balancing is improved. In Figure \ref{fig:res-failed} the agent failed to balance the initial level with $b_0=0$. The balancing process terminated after the maximum of permitted changes has been reached. In the end $b$ is still 0.

\subsection{Impact of tiles on balancing}
\label{sec:impact}
The analysis of the actual swaps a model made gives insights on its behavior. Thus, we showed the model could improve the balancing state of the given levels, we can further argue that swaps made by the model have impact on the balancing. In reverse, we can conclude from this behavior which tiles in the game have most impact on balancing.

\begin{figure}[!t]
  \centering
  \includegraphics[width=3.5in]{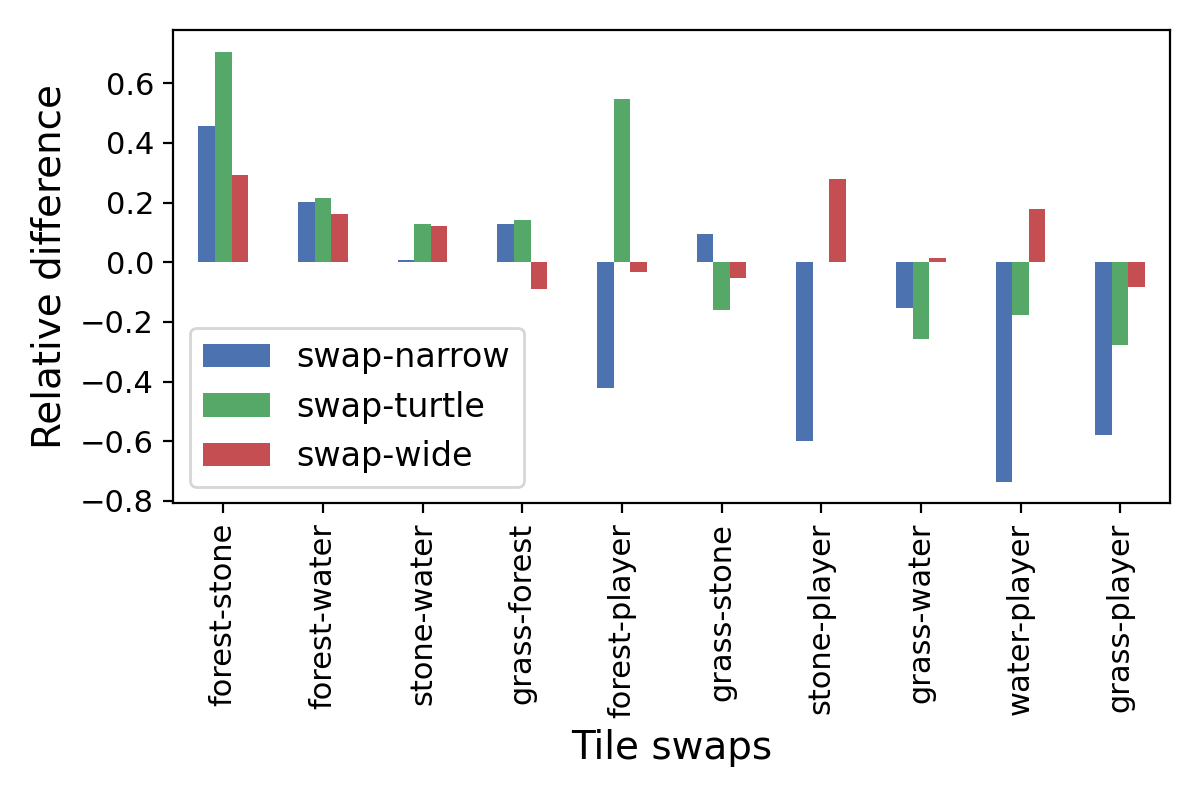}
  \caption{Comparing the swapped tiles by the model per representation on the generated 1000 levels. The comparison is made in relation to the inverse tile type distribution of all levels. This shows the difference to random swapping.}
  \label{fig:swaps}
\end{figure}

For comparison, we calculate per swap pair the relative differences of frequencies of random swaps with the ones from the models. Therefore, these frequencies are factorized with the inverse probabilities of total tile occurrences in the data set. Due to swapping the same tile types is prohibited by the representations, ten different combinations are possible. This is depicted in Figure~\ref{fig:swaps}.
The figure shows the three representations have different behaviors; however, they agree in particular points. Most impact has the swapping of forest with stone tiles, second the swap of forest with water tiles. This makes sense in the context of the resource gathering win condition. Relocating the resource tiles forest or water will likely affect the balancing state thus, they are included in the win condition.
A stone field can be used to block the way for players. Due to this, the swapping of forest with stone tiles has an additional powerful impact.
Surprisingly swapping players' spawn positions is not a favored action in all cases.

\section{Discussion and limitations}
\label{sec:discussion}
Our experiments showed promising results within fewer training steps compared to the original PCGRL while being more robust to ensure playability. There are, however, several things that need to be discussed.

The reward of the balancing agent represents the balancing state of multiple simulations on the current level state. By design, this is done with the use of the information of which player actually won per simulation pass. Using game specific information such as e.g., the health state of players inside the reward function potentially might improve the results. This is, however, not desired since including domain specific information creates dependencies on the distinct game. In this case, a special reward function must be developed for each game. Furthermore, including too much or the wrong information could bias the reward resulting in poor performance or unwanted behavior of the model. As a consequence, the model would not learn what \emph{really} influences balancing, or it might exploit unforeseen loopholes.

Splitting up the level generation and balancing process into separated units yielded much better results than doing it in the same step as in the original PCGRL. 
The plain PCGRL might be able to achieve comparable results when trained significantly longer, however, the here proposed architecture converges faster in less updates of the policy. This gives evidence that our architecture is more efficient by subdividing the problem into two simpler problems. In combination with the swapping representations, the balancing process can fully focus on balancing. We compared three adapted representations from the original PCGRL and evaluated them regarding performance. All of them can improve the balancing state, whereas swap-turtle yielded slightly worse results than the other two. Thus, swap-narrow, and swap-wide perform comparably, we recommend using the narrow representation due to its small action space. It is, furthermore, independent from the grid size.

Since we showed, that the model improves a level's balancing significantly, it is possible to draw conclusions about which tile types have most impact on balancing regarding their swap-frequencies.
For this domain, the swapping of resource tiles (e.g., forest) with blocking elements (stone) had the greatest impact.

By simulating the game $n$-times with player agents the balancing state is evaluated. This metric is of course dependent on exactly these types of players. When using players with e.g., different skill or type the balancing would be different. That is a limitation, however, being also an advantage. By using different types of player agents, the game level could be balanced to compensate skill differences of the players by solely adjusting the level not the players themselves. This is of high interest when balancing levels for different player types such as a mage and a fighter. It can be applied in e.g.; role play games where gear levels are an indicator of the character strength. So, players can use their long-farmed equipment in a competitive setting and the game could be balanced through the environment only. In future research, our work is to be extended towards this. The controllable content generators~\cite{earle_learning_2021} might be an applicable solution in that context.

An issue of the proposed architecture is a computational effort in the training of the model which lies in the simulation step for rewarding. For each swap the agent takes on the level the simulation step is run. In addition, we figured out that this reward is mostly sparse in training. That indicates, only a small sub sample of the action space are actually actions which influence the balancing. Especially in the beginning of the training, this must be learnt by the model first.
To fasten the training process, we think of methods to reduce the computational cost. That would be also of high interest for the application in more complex environments.
One solution might be to reward the agent after several time steps only or even use sparse rewards. Despite the learning process would then be harder for the agent, simulation steps for the reward are omitted. Thus, the training process is sped up and the agent can explore faster.
Another approach could be the usage of a reward model. Therefore, a model is trained to predict the balancing state of a level. This would speed up training a lot, however, the model's accuracy must be high enough to give suitable rewards to ensure a correct training.

\section{Conclusion and Future Work}
\label{sec:conclusion}
In this paper, we proposed a new method to balance game levels with reinforcement learning for tile-based games.
Therefore, we introduced a novel swap-based representation family, where agents can swap the locations of tiles in a given level.
Our architecture benefits from separating the level generation and balancing process into two subsequent processes. 
While training a balancing model, we reward the model with the actual balancing state, which is simulated by player agents. The results show that this method balances levels in 48.1\% of cases and improves the balancing state of levels in 64.1\%. Compared to the original PCGRL method (30.4\%) on generated levels of the NMMO environment, our approach is easier to learn and significantly improves the results. 
Furthermore, by analyzing which tiles are swapped frequently, information can be inferred which tile types influence the balancing most. This yields additional empirical proof for game designers how the game system itself actually behaves.

For future work, we want to test our method with increased player sizes. Additionally, the balancing for players of different skill by only altering the game level is of interest. This opens new perspectives to create levels when e.g., adults play vs. children, but also for the competitive play of players with different e.g., gear levels. Moreover, we think that this method is applicable not only to games, but also to other domains as well. In subsequent research we are interested in the application to city planning for fair infrastructure distribution.

\bibliographystyle{unsrt}
\bibliography{cog.bib}

\end{document}